\documentclass[letterpaper, 10 pt, conference]{ieeeconf}  

\IEEEoverridecommandlockouts                              

\usepackage{graphics} 
\usepackage{epsfig} 
\usepackage{mathptmx} 
\usepackage{times} 
\usepackage{amsmath} 
\usepackage{amssymb}  
\usepackage{caption}
\usepackage{cite}
\usepackage{booktabs}
\usepackage{makecell}
\usepackage{multirow}
\usepackage{etoolbox}   
\usepackage{bbding} 
\usepackage{fontawesome5}
\title{\LARGE \bf
FAR-Dex: Few-shot Data Augmentation and Adaptive Residual Policy Refinement for Dexterous Manipulation
}

\author{Yushan Bai$^{1,*}$, Fulin Chen$^{2,*}$, Hongzheng Sun$^{1,*}$, Yuchuang Tong$^{1,\dagger}$, En Li$^{1,3}$, Zhengtao Zhang$^{1,3,\dagger}$%
	\thanks{This work was supported by the National Natural Science Foundation of China (62303457), the Hebei Provincial Science and Technology Plan Project (25241802D), and the New Generation Artificial Intelligence-National Science and Technology Major Project (2025ZD0122900).}%
	\thanks{$^{1}$Institute of Automation, Chinese Academy of Sciences, Beijing 100190, China, also with the School of Artificial Intelligence, University of Chinese Academy of Science, Beijing 100049, China, also with the CAS Engineering Laboratory for Intelligent Industrial Vision, Beijing 100190, China. E-mail: \{baiyushan2023, sunhongzheng2024, yuchuang.tong, en.li, zhengtao.zhang\}@ia.ac.cn}%
	\thanks{$^{2}$School of Mechanical and Automotive Engineering, Shanghai University of Engineering Science, Shanghai 201620,  E-mail: chenfulin@sues.edu.cn}%
	\thanks{$^{3}$Beijing Zhongke Huiling Robot Technology Co., Beijing 100192, China.}%
	\thanks{* Equal contribution. $\quad$ $^{\dagger}$ Corresponding authors.}%
}

\begin{document}

\maketitle
\thispagestyle{empty}
\pagestyle{empty}

\begin{abstract}

Achieving human-like dexterous manipulation through the collaboration of multi-fingered hands with robotic arms remains a longstanding challenge in robotics, primarily due to the scarcity of high-quality demonstrations and the complexity of high-dimensional action spaces. To address these challenges, we propose FAR-Dex, a hierarchical framework that integrates few-shot data augmentation with adaptive residual refinement to enable robust and precise arm–hand coordination in dexterous tasks. First, FAR-DexGen leverages the IsaacLab simulator to generate diverse and physically-constrained trajectories from a few demonstrations, providing a data foundation for policy training. Second, FAR-DexRes introduces an adaptive residual module that refines policies by combining multi-step trajectory segments with observation features, thereby enhancing accuracy and robustness in manipulation scenarios. Experiments in both simulation and real-world demonstrate that FAR-Dex improves data quality by 13.4\% and task success rates by 7\% over state-of-the-art methods. It further achieves over 80\% success in real-world tasks, enabling fine-grained dexterous manipulation with strong positional generalization.
\end{abstract}

\section{Introduction}
In recent years, imitation learning\cite{zhao2023learning, chi2025diffusion, ze2024dp3} has emerged as the dominant approach for dexterous manipulation, as it enables robots to acquire complex skills directly from human demonstrations. Most existing methods\cite{qin2022dexmv, mandi2025dexmachina} focus on either the robotic arm or the dexterous hand in isolation. However, achieving human-like dexterity requires the seamless integration of multi-fingered hands with robotic arms, yet such coordination substantially increases the dimension of action space, making unified control particularly challenging\cite{chen2025object,fu2025cordvip}. Moreover, the scarcity of high-quality demonstrations and the lack of fine-grained hand–object interaction details\cite{li2023gendexgrasp,zhao2024graingrasp} further hinder real-world deployment, underscoring that both the quantity and quality of data are essential for training robust and generalizable policies. Therefore, addressing these challenges calls for the development of new frameworks that can leverage limited demonstrations, enhance system scalability, and enable reliable arm–hand coordination in fine-grained manipulation tasks.

\begin{figure}[htbp]
	\centering
	\includegraphics[width=0.46\textwidth]{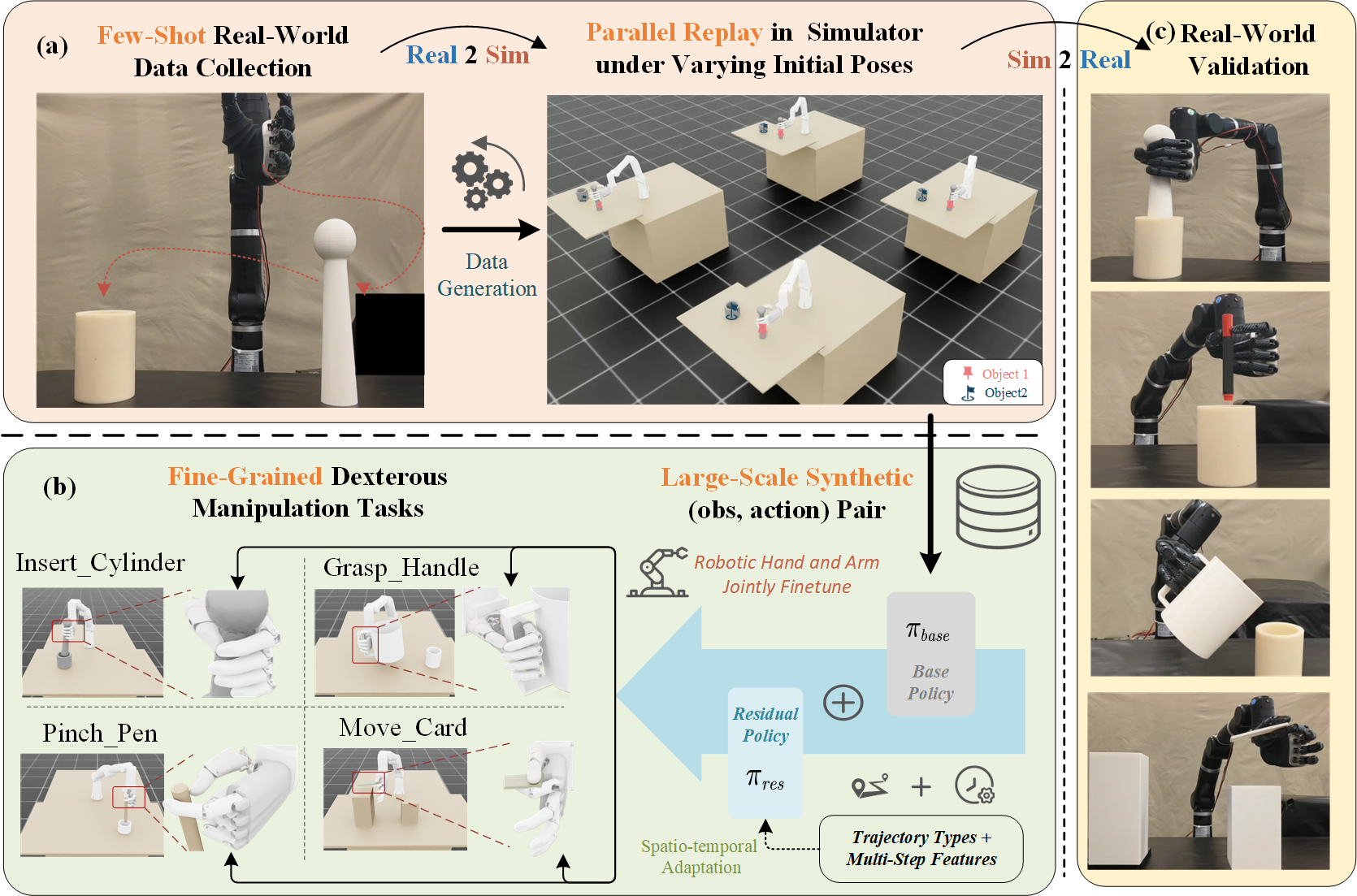}
	\captionsetup{font=small,skip=8pt,belowskip=0pt}
	\caption{Overview of the proposed FAR-Dex. (a) FAR-DexGen: demonstration trajectories are decomposed and transformed to generate large-scale synthetic data in the simulator. (b) FAR-DexRes: spatio-temporal residual fine-tuning is applied to four dexterous manipulation tasks. (c) Real-world validation: the trained policy is directly deployed in physical environments.}
	\label{fig:1}
\end{figure}

To alleviate the scarcity of demonstration data, prior studies\cite{mandlekar2023mimicgen, jiang2025dexmimicgen, garrett2025skillmimicgen} have developed automated data augmentation systems in simulation, where original trajectories are transformed and extended to synthesize diverse datasets. However, such methods often suffer from significant performance degradation when transferred to real-world environments. To mitigate this gap, DemoGen\cite{xue2025demogen} incorporates real-world point clouds to directly generate demonstrations in physical settings, thereby reducing domain discrepancies. Nonetheless, its insufficient modeling of arm-hand coordination limits its effectiveness in fine-grained manipulation tasks.

Beyond data augmentation, another line of research seeks to improve the fine-grained performance of dexterous manipulation through residual policy refinement. The central idea is to employ online reinforcement learning (RL) to refine a base policy via residual corrections\cite{yuan2024policydecorator}, thereby enhancing adaptability to complex environments. Some studies have extended this approach to dexterous manipulation\cite{li2025maniptrans, mandi2025dexmachina,ankile2025imitation}, where dynamic error compensation during execution improves stability. However, the lack of explicit spatio-temporal modeling limits the precision and robustness of these methods in long-horizon manipulation tasks.

To address these challenges, we propose FAR-Dex—a hierarchical framework (Fig. 1) that integrates few-shot data augmentation with adaptive residual refinement, achieving both strong generalization across tasks and precise execution. Unlike existing approaches\cite{xue2025demogen,yuan2024policydecorator}, FAR-Dex not only synthesizes diverse, detailed 3D data from limited demonstrations to mitigate the scarcity of high-quality samples, but also performs residual fine-tuning by incorporating multi-step trajectory segments and observation features, thereby enhancing the precision and robustness of arm–hand coordination. Extensive experiments in both simulation and real-world settings demonstrate that, compared to state-of-the-art methods, FAR-Dex improves data quality and task success rates by 13.4\% and 7\%, respectively, and achieves over 80\% success in real-world execution, effectively bridging the gap between limited demonstrations and practical deployment.

The main contributions of this paper are as follows:
\begin{itemize}
\item We propose FAR-Dex, a hierarchical framework that integrates few-shot data augmentation with adaptive residual refinement, enabling coordinated arm–hand dexterous manipulation with robustness and high precision, even from limited demonstrations.
\item We propose a data generation system that mitigates the scarcity of fine-grained hand–object interaction data by synthesizing diverse, physically-constrained trajectories, thus improving efficiency and scalability.
\item We design an adaptive residual refinement module that incorporates spatio-temporal adaptive weights to dynamically regulate the residual corrections to the base policy, achieving more fine-grained and robust arm–hand coordinated control. 
\end{itemize}

\section{Related Works}

\subsection{Data Generation based on few demonstration}

In robotic manipulation, data generation has emerged as an effective means to reduce collection costs and enhance policy generalization. Some approaches\cite{wang2024cyberdemo,kanehira2025rl,dasari2023learning} leverage reinforcement learning to increase data diversity in simulation, but RL-generated behaviors often deviate from natural human actions. The MimicGen family\cite{mandlekar2023mimicgen, garrett2025skillmimicgen, jiang2025dexmimicgen} synthesize human-like data through trajectory slicing in simulation, yet they lack fine-grained 3D interaction details, resulting in significant errors when transferred to reality. Other studies\cite{yu2025real2render2real,ren2025learning,wu2025rl} attempt to use 3D reconstruction for high-fidelity rendering to reduce the sim-to-real gap; however, these methods are slow and limited to static grasping tasks, restricting their training applicability. DemoGen\cite{xue2025demogen} introduces point cloud data to accelerate generation and enrich contact information, but the reliance on point cloud stitching leads to visual mismatches and the absence of dynamic modeling, making it unsuitable for fine-grained manipulation.

To overcome these limitations, we leverage trajectory segmentation \cite{garrett2025skillmimicgen} and 3D recombination \cite{xue2025demogen} to automate data collection via demonstration replay in IsaacLab\cite{mittal2025isaac}. This approach ensures high data quality for policy training while effectively minimizing domain gaps.

\subsection{Residual Policy Learning}

In robotic manipulation, residual policies\cite{johannink2019residual} are often employed to refine base policies online and mitigate error accumulation during inference. Policy Decorator\cite{yuan2024policydecorator} achieves rapid fine-tuning by adding lightweight residuals at the output layer, but its application remains confined to simulation and neglects the varying residual requirements across different task phases. Similar to this idea, ResiP\cite{ankile2025imitation} deployed residual policies to high-precision assembly tasks by applying residual corrections on baseline actions; however, its use of a single scaling factor across the entire action space limits applicability in high-DoF dexterous manipulation. Further, ManipTrans\cite{li2025maniptrans} applied the residual paradigm to multi-fingered dexterous hand control, yet its heavy reliance on high-quality demonstrations and strict alignment restricts its generalization in complex scenarios.

To address this, we develop an adaptive residual refinement module that utilizes trajectory and temporal context for phase-sensitive weighting. These dimensionally aligned weights facilitate precise, fine-grained coordination between the robotic arm and dexterous hand.

\section{Methods}

\subsection{Problem Formulation} 
The coordination of multi-fingered dexterous hands with robotic arms faces two key challenges: (i) limited human demonstrations are both scarce and lacking in detailed 3D interaction information; and (ii) the high-dimensional action space makes precise control in long-horizon tasks highly challenging. To formulate these issues, we first collect a few human demonstrations with a trajectory length of $N$:
\begin{equation}
	D_h=\{d_t=(o_t,a_t)\}^{N}_{t=1}|c_h
	\label{eq1}
\end{equation}
where $o_t=(o_t^{\text{pcd}},o_t^{\text{arm}},o_t^{\text{hand}})$  includes the scene point cloud and arm–hand states, $a_t=(a_t^{\text{arm}},a_t^{\text{hand}})$  represents control commands, and $c_h=\{ \mathbf{T}^{O_1}_h, \dots, \mathbf{T}^{O_K}_h \}$ denotes the initial poses of $K$ objects.  The aim is to learn a joint policy $\pi:\mathit{O}\to\mathit{A}$ with robust and precise arm–hand coordination across tasks.

To this end, we propose FAR-Dex, a hierarchical framework. As illustrated in Fig. 2, FAR-DexGen expands the limited demonstrations into a large-scale dataset $D_g$ by applying trajectory segmentation and simulation-based augmentation, synthesizing training data that preserve physical constrain and rich spatial information. Building on this, FAR-DexRes introduces a residual refinement policy $\pi_{\text{res}}$, which employs an adaptive weighting mechanism to dynamically adjust actions based on spatio-temporal states, thereby achieving finer arm–hand control in long-horizon tasks.

\begin{figure*}[htbp]
	\centering
	\includegraphics[width=0.92\textwidth]{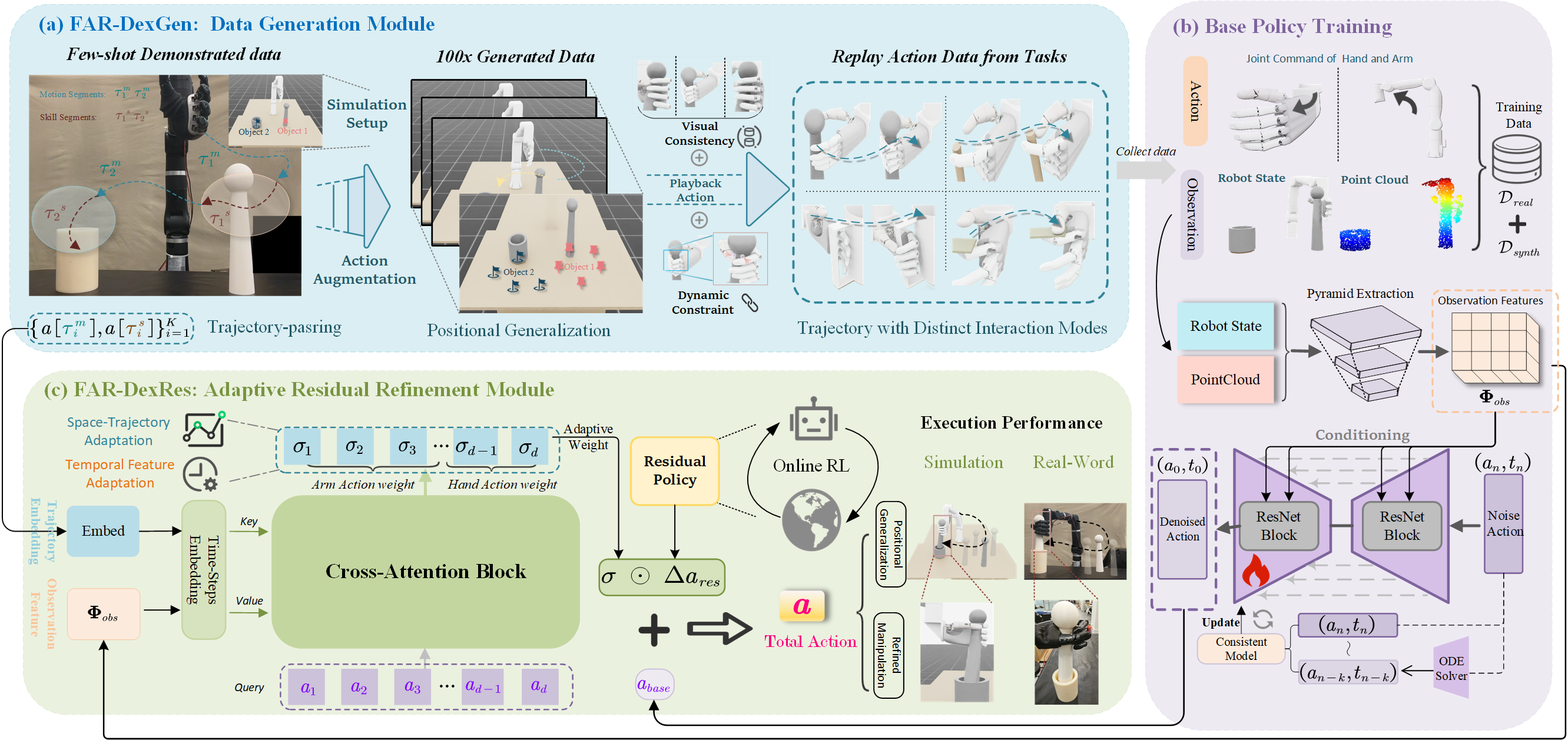}
	\captionsetup[figure]{font=small,skip=4pt,belowskip=0pt}
	\caption{FAR-Dex framework pipeline. (a) Few demonstrations $D_h$ are segmented and augmented via spatial transformations in IsaacLab to form a large-scale dataset $D_g$. (b) The combined data are encoded with a pyramid convolutional network, where a consistency model distills the denoising network of $\pi_{\text{base}}$ to accelerate inference (c) Building on $a_{\text{base}}$ from $\pi_{\text{base}}$, FAR-DexRes integrates multi-step trajectory embedding and observation features to generate adaptive weights $\sigma$ for spatio-temporal residual refinement in dexterous manipulation.
	}
	\label{fig:2}
\end{figure*}

\subsection{FAR-DexGen: Data Generation Module}
To enable scalable dexterous manipulation under limited demonstrations, FAR-DexGen expands few human demonstrations into a large-scale set of physically feasible trajectories while preserving visual consistency and fine-grained interaction details. This module primarily addresses two challenges: (i) capturing hand–object spatial interaction information, and (ii) reducing the sim-to-real transfer gap.
\subsubsection{Action Sequence Parsing}
In dexterous manipulation, raw demonstrations are often unsuitable for direct policy learning due to noise and uncontrolled environments. To reduce real-to-sim transfer errors, we adjust the physical parameters of the simulation environment to ensure stable replay of the action sequences $A_h=\{a_1,a_2,\dots,a_{N} \mid c_h\}$ from $D_h$. Since treating the entire trajectory as unified whole makes it difficult to balance global generalization with local precision, we draw inspiration from prior works\cite{xue2025demogen,garrett2025skillmimicgen} and introduce a trajectory segmentation mechanism, dividing trajectories into motion segments and skill segments.

\begin{itemize}
	\item  \textit{Motion segments}: The process in which the robot adjusts its pose and approaches the object from free space.
	\item \textit{Skill segments}: The execution of fine-grained actions, including pre-grasping, contact, and manipulation.
\end{itemize}

In practice, we compute the distance between the geometric center of the object point cloud and the dexterous hand. When this distance falls below a predefined threshold, the corresponding trajectory segment are marked as \textbf{Skill segments}. Trajectory segments located between adjacent skill segments are classified as \textbf{Motion segments}. Consequently, the action sequence is parsed into an alternating structure of motion and skill phases:
\begingroup
\setlength{\abovedisplayskip}{6pt}    
\setlength{\belowdisplayskip}{6pt}    
\setlength{\abovedisplayshortskip}{3pt}
\setlength{\belowdisplayshortskip}{3pt}
\begin{equation}
	A'_{h} = \{a[\tau_1^m], a[\tau_1^s], \dots, a[\tau_K^m],a[\tau_K^s]|c_0\}
	\label{eq2}
\end{equation}
\endgroup
where $\tau = \{t_{\text{start}},t_{\text{start}+1}, \dots, t_{\text{end}-1},t_{\text{end}}\}$ represents a set of multiple time steps, $\tau_k^m$ and $\tau^s_k$ correspond to the motion segment of approaching the 
$k$-th object and the skill segment of interacting with the $k$-th object, respectively. This parsing mechanism enables simultaneous preservation of global spatial generalization and local interaction precision.

\subsubsection{Action Synthesis}
After parsing, FAR-DexGen generates diverse trajectories by varying the initial object poses while maintaining a fixed initial robot configuration. Inspired by DemoGen \cite{xue2025demogen}, such pose variations not only alter the end-effector configurations within skill segments but also propagate to adjacent motion segments, thereby constructing a wide range of differentiated action trajectories within the robot’s reachable workspace.

To ensure physical feasibility, arm joint angles are mapped to end-effector poses via forward kinematics: $P_t^{\text{arm}}=FK(J_t^{\text{arm}})$. The corresponding object pose transformation $\Delta c$ (from initial $c_h$ to target $c_i$) is uniformly sampled at 5 cm intervals across all tasks. This systematic sampling ensures unbiased spatial coverage during data generation,enhancing the framework’s generalization across the workspace.

Fig. 3 illustrates this process and the resulting generated trajectories. The transformation directly affects both arm and hand execution. As the two components serve distinct roles across trajectory segments, we model them separately during synthesis, as detailed below:

\textbf{Robotic Arm:} For skill segment $\tau^s_k$, the end-effector poses $P_t^{\text{arm}}[\tau^s_k]$ are adjusted by $\Delta c$, and converted into the corresponding joint angles $\hat{a}^{\text{arm}}[\tau^s_k]$ are obtained through inverse kinematics:
\begin{equation}
	\hat{a}^{\text{arm}}[\tau^s_k] = \text{IK}(P^{\text{arm}}[\tau^s_k] \cdot \Delta c)
	\label{eq3}
\end{equation}

For motion segments $\tau^m_k$, motion planning\cite{dalal2023imitating,mandlekar2023human} is used to smoothly connect the end-effector pose of the previous skill segment and the initial pose of the next one, which are then converted into joint angles: 
\begin{equation}
	\hat{a}^{\text{arm}}[\tau^m_k] = \text{IK}( \text{MotionPlan}( \hat{P}^{\text{arm}}\tau^s_{k}[-1],  \hat{P}^{\text{arm}}\tau^s_{k+1}[0] ))
	\label{eq4}
\end{equation}

\begin{figure}[]
	\centering
	\includegraphics[width=0.46\textwidth]{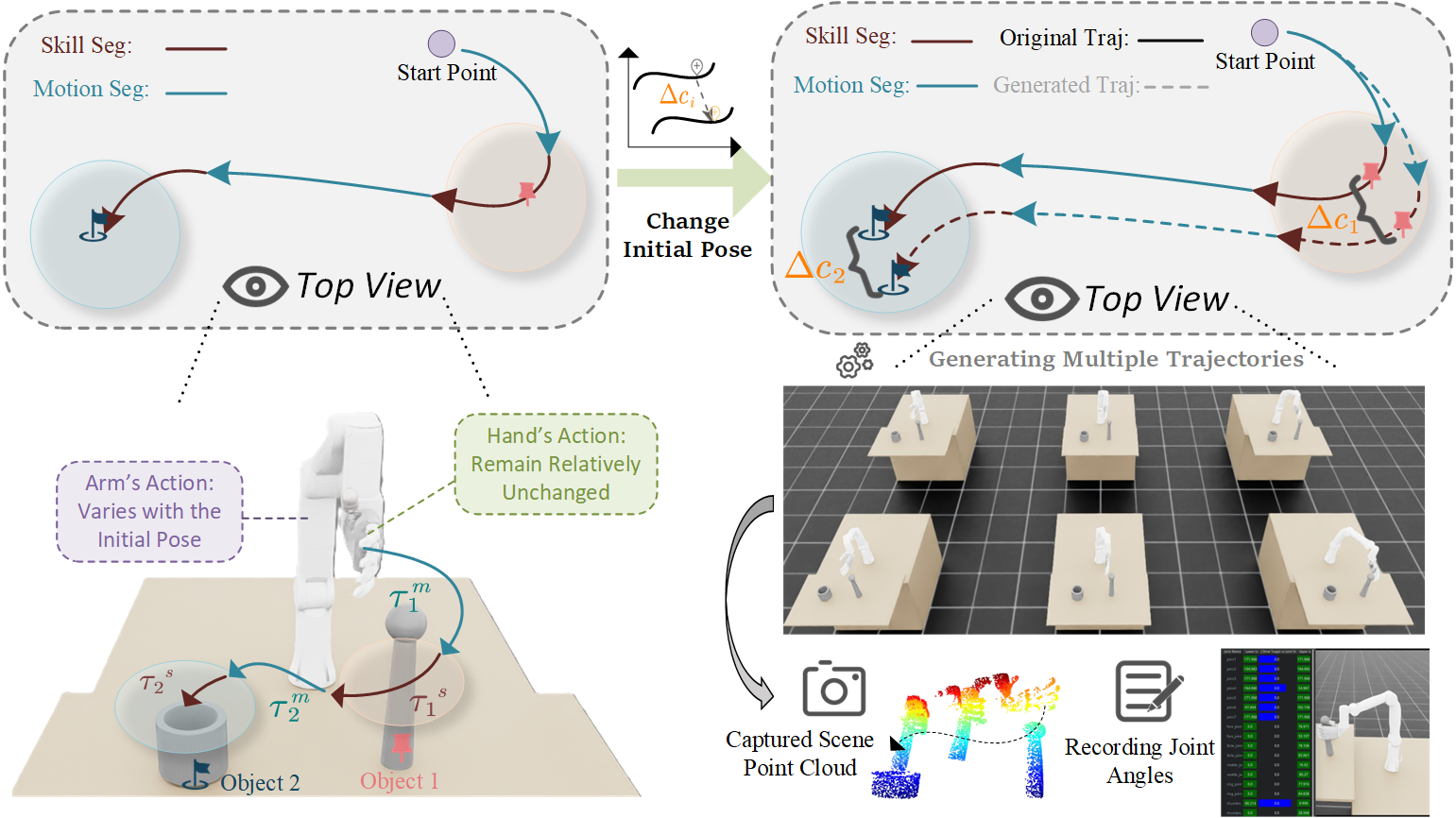}
	\captionsetup{font=small,skip=8pt,belowskip=0pt}
	\caption{Data generation pipeline. Demonstration trajectories are illustrated using a top view, where transformations of the initial object pose $\Delta c_i$ induce corresponding variations across different trajectory segments. These trajectories are then deployed in the simulation environment to collect data.}
	\label{fig:3}
\end{figure}

\textbf{Dexterous Hand:} In contrast, the dexterous hand primarily handles contact and manipulation and is less sensitive to spatial perturbations. Thus, its actions remain identical to those in the original demonstrations:
\begin{equation}
	\hat{a}^{\text{hand}}[i] = J^{\text{hand}}[i], \quad i \in \{ \tau^m_1,\tau^s_1, \dots, \tau^m_K,\tau^s_K \}
	\label{eq5}
\end{equation}

Finally, we combine the transformations of the robotic arm and dexterous hand to obtain the complete synthesized action sequence $\hat{A}_h^i$ under the new initial configuration $c_i$:
\begin{equation}
	\hat{A}_h^i = \{(\hat{a}^\text{arm}[\tau_1^m],\hat{a}^\text{hand}[\tau_1^m]), \dots, (\hat{a}^\text{arm}[\tau_K^s],\hat{a}^\text{hand}[\tau_K^s])|c_i\}
	\label{eq6}
\end{equation}
where both $\hat{a}^\text{arm}$ and $\hat{a}^\text{hand}$ are represented by joint angles. Using joint-angle representation preserves true robot motion states, enabling high-fidelity simulation replay and direct deployment in training.

\subsubsection{Data Collection in Simulation Environment}
After obtaining the synthetic action sequences $\hat{A}^i_h$ under different configurations $c_i$ from (6), we deploy them in the IsaacLab simulator\cite{mittal2025isaac} to replay action trajectories and collect observation–action pairs $D_g=\{d_t=(o_t,a_t)\}^{N}_{t=1}|c_i$. Unlike DemoGen’s offline stitching \cite{xue2025demogen}, our online synthesis captures dynamic spatial interactions, including precise finger-object alignment and relative pose variations. This physics-aware process implicitly encodes critical contact constraints and ensures viewpoint consistency, providing the policy with a robust representation of the hand-object interaction state.

To further reduce sim-to-real transfer errors, point clouds are transformed into the robot arm coordinate frame and robustly aligned through cropping, clustering, and sampling\cite{qin2023dexpoint}. Domain randomization is applied by injecting Gaussian noise into the observations to improve robustness against sensing errors and environmental disturbances. In addition, the built-in collision detection of IsaacLab ensures the physical feasibility of the trajectories. 

Finally, task-specific functions are employed to filter out failed trajectories, resulting in a training dataset $\mathit{D}=D_h\cup D_g$ that preserves both visual consistency and dynamic feasibility, providing a reliable foundation for policy training.

\subsection{FAR-DexRes: Residual Refinement Module}
To enhance online dexterous manipulation performance, FAR-DexRes incorporates a residual refinement module that simultaneously improves inference efficiency and fine-grained control. Its design centers on two components: (i) training the base policy with a consistency model to reduce inference latency, and (ii) an adaptive residual policy for dynamic error correction in long-horizon tasks.

\subsubsection{Base Policy Training with Consistency Models}
Through FAR-DexGen, we construct a training dataset $\mathit{D}=D_h\cup D_g$ that ensures both spatial consistency and dynamic feasibility, and use it to train the base policy $\pi_{\text{base}}$ within the DP3\cite{ze2024dp3} framework. While DP3 is capable of generating high-quality action sequences, it requires $\mathit{O}(K)$ sampling steps when processing high-dimensional point clouds and complex interactions, resulting in excessive inference overhead that often leads to latency and instability in long-horizon arm–hand coordination tasks.

To mitigate this, FAR-DexRes incorporates a consistency model\cite{song2023consistency,luo2023latent} inspired by ManiCM\cite{lu2024manicm}, distilling the multi-step DP3 denoising into a single-step predictor to enforce consistency across noise scales. Given a clean action sequence $a \in \mathbb{R}^{T \times d}$, a noisy sample $\tilde{a}_n$ at scale $n$ is obtained following DDIM\cite{song2020denoising}, and the teacher model denoises it through a $K$-step ODE solver $\Phi(\cdot)$ to produce $\tilde{a}_{n-k}$. 

Since $\tilde{a}_n$ and $\tilde{a}_{n-k}$ are consistent on the low-dimensional action manifold, the consistency network $g_\theta$ is distilled to minimize cross-scale prediction discrepancies by learning from the teacher’s noise prediction model $g_{\theta^-}$ with the following loss:
\begingroup
\setlength{\abovedisplayskip}{6pt}    
\setlength{\belowdisplayskip}{6pt}    
\setlength{\abovedisplayshortskip}{3pt}
\setlength{\belowdisplayshortskip}{3pt}
\begin{equation}
	\mathcal{L}_{\text{CM}}(\theta) = \mathbb{E}_{a,t}\| g_{\theta}(\tilde{a}_{n-k} ,t_{n-k}) - g_{\theta^{-}}(\tilde{a}_n ,t_n) \|^2_2
	\label{eq7}
\end{equation}
\endgroup

This enforces consistent predictions across noise levels and constrains the action distribution on the low-dimensional manifold. At inference, the model only requires the observation $o_t$ to generate a high-quality action $\hat{a}_s$ in a single step:
\begingroup
\setlength{\abovedisplayskip}{6pt}    
\setlength{\belowdisplayskip}{6pt}    
\setlength{\abovedisplayshortskip}{3pt}
\setlength{\belowdisplayshortskip}{3pt}
\begin{equation}
	\hat{a}_s = g_{\theta}(f_{\phi}(o_t),t_s\approx 0)
	\label{eq8}
\end{equation}
\endgroup
where $f_\phi(\cdot)$ is the observation encoder. To enhance point cloud representation, we replace the MLP in DP3 with a four-stage recursive PointNet encoder. Each stage integrates PointWise convolutions with global-local feature aggregation via ReLU activation, distilling geometric information into a 128-dimensional embedding. This hierarchical design captures both local geometric details and global context, enabling efficient real-time dexterous manipulation with reduced inference latency.

\subsubsection{Adaptive Residual Policy Refinement}
Although the consistency-distilled $\pi_{\text{base}}$ from (\ref{eq7}) can rapidly generate control actions $a_{\text{base}}$ in most cases, its training relies solely on offline data and lacks critical contact details from online interaction. During real execution, this deficiency often prevents timely correction of out-of-distribution states, leading to the gradual accumulation of prediction errors. To address this issue, we introduce a residual policy $\pi_{\text{res}}$, which dynamically adjusts control signals through online reinforcement learning. This design preserves global smoothness while enabling fine-grained corrections during contact phases.

\begin{figure}[]
	\centering
	\includegraphics[width=0.37\textwidth]{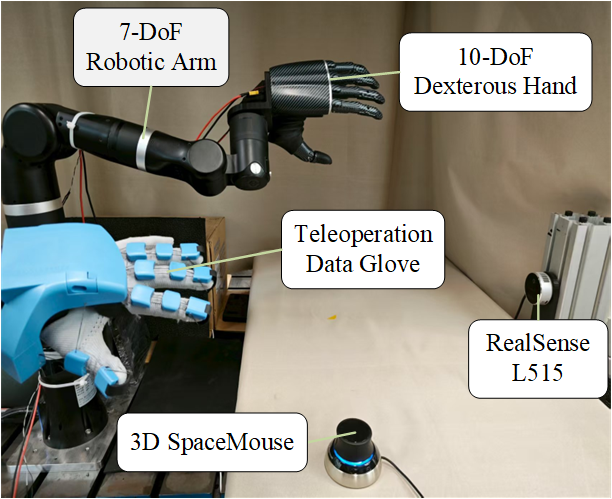}
	\captionsetup{font=small,skip=8pt,belowskip=0pt}
	\caption{Demonstration Data Collection System in the Real World.}
	\label{fig:4}
\end{figure}

To ensure the residual policy adaptively corrects phase-specific errors, we design a cross-attention weighting network. Specifically, we utilize a temporal horizon of $H$ steps to capture multi-step features, where the corresponding $H$-step observation sequence $\Phi_{\text{obs}}$ and trajectory embeddings $E_{\text{traj}}$ serve as the key $K$ and value $V$, respectively. As shown in Fig. 2-c, the base action $a_{\text{base},t}$ is projected as the query $Q$ to interact with these features. To preserve temporal context, learnable positional embeddings are added before applying Layer Normalization. The process yields element-wise weights $\sigma_t$ via a Sigmoid-activated linear layer:

\begingroup
\setlength{\abovedisplayskip}{0pt}    
\setlength{\belowdisplayskip}{4pt}    
\setlength{\abovedisplayshortskip}{1pt}
\setlength{\belowdisplayshortskip}{3pt}
\begin{equation}
	\sigma_t = \text{MultiAttn}(Q=a_{\text{base,t}},K,V=\{ E_{\text{traj}},  \Phi_{\text{obs}} \})
	\label{eq9}
\end{equation} 
where $\sigma_t \in \mathbb{R}^{1\times d}$ from (\ref{eq9}) is strictly aligned with the action space, enabling each action component to receive an independent residual weight modulation. The total action $a_{\text{total,t}}$ can thus be expressed as:

\begingroup
\setlength{\abovedisplayskip}{6pt}    
\setlength{\belowdisplayskip}{6pt}    
\setlength{\abovedisplayshortskip}{3pt}
\setlength{\belowdisplayshortskip}{3pt}
\begin{equation}
	a_{\text{total,t}} = a_{\text{base},t} + \sigma_t \odot a_{\text{res},t}
	\label{eq10}
\end{equation} 
\endgroup

Trained via PPO\cite{schulman2017proximal}, $\pi_{\text{res}}$ employs a warm-start strategy\cite{yuan2024policydecorator} to ensure stability. By generating adaptive residual action $a_{\text{res},t}$ to refine the base action $a_{\text{base},t}$ from $\pi_{\text{base}}$, FAR-DexRes significantly improves dexterous manipulation in complex, long-horizon tasks and real-world scenarios.

\section{Experiments}
We conduct comprehensive experiments in both simulation and real-world to investigate the following questions:
\begin{itemize}
\item What advantages does our method offer in data generation compared to existing approaches?(Sec.IV-B)
\item How does the adaptive residual policy improve performance in dexterous manipulation tasks? (Sec.IV-C,D)
\item How are adaptive residual weights $\sigma_t$ distributed in arm–hand coordinated control? (Sec.IV-E)
\item As a real2sim2real framework, how do the two transfer stages affect performance in real tasks? (Sec.IV-F)
\end{itemize}

\subsection{Experimental Setup}
\subsubsection{Robot Data Collection System}
As shown in Fig. 4, the experimental setup consists of a 7-DoF Realman Gen72 robotic arm, a 10-DoF Casbot P0S dexterous hand, and an Intel RealSense L515 RGB-D camera for capturing scene point cloud. Demonstrations are collected using a 3D SpaceMouse for precise end-effector control, while finger joint angles measured by a teleoperation data glove are mapped to the dexterous hand for natural teleoperation. For each task, two expert demonstrations are recorded at 20 Hz to ensure the capture of fine-grained motion details.

\subsubsection{Parameter settings}
In the training configuration, the trajectory segmentation threshold is set to 0.1m to distinguish motion and skill segments. The temporal length $H$ is fixed at 8 steps to capture multi-step features. For RL fine-tuning, all tasks employ sparse rewards defined by target proximity, object lifting height, and hand–object distance. The cross-attention block is implemented with a single layer. All network training and inference are executed on an Intel i9-14700K CPU and a NVIDIA RTX 4090D GPU.

\subsubsection{Task Design}
To comprehensively evaluate performance in complex contact and fine-grained manipulation, we design four dexterous tasks, each capped at 1500 steps. As shown in Fig.1-b, these tasks require the dexterous hand to complete precise interactions under different contact modes:

\begin{table}
	\renewcommand{\arraystretch}{1.2}
	\setlength{\tabcolsep}{4pt} 
	\centering
	\captionsetup{font=small}
	\caption{Comparison of different data generation methods}
	\label{tab1}
	\begin{tabular}{l|c c}
		\toprule[1.1pt]
		Method & Generation Time ($T_g$) & Generation Quality ($\eta_g$) \\ \midrule
		MimiGen\cite{mandlekar2023mimicgen} & \textbf{8.3ms}    & 68.3\%  \\
		DemoGen\cite{xue2025demogen}        & 9.1ms & 74.5\%  \\
		\textbf{FAR-DexGen (Ours)}          & 10.3ms & \textbf{87.9\%} \\ 
		\bottomrule[1.1pt]
	\end{tabular}
\end{table}

\begin{itemize}
	\item \textbf{Insert Cylinder}: Hold a cylindrical object and insert it into a narrow hole, which requires spatial alignment.
	\item \textbf{Pinch Pen}: Retrieve a pen from a pen holder using a thumb–index pinch, which is sensitive to pose errors. 
	\item \textbf{Grasp Handle}: Grasp and slightly rotate a kettle handle, which demands continuous finger joint adjustment.
	\item \textbf{Move card}: Move a card to another tabletop using a lateral thumb–index pinch, needing precise coordination.
\end{itemize}

\begin{table*}[]
	\centering
	\renewcommand{\arraystretch}{1.2}
	\captionsetup{font=small}
	\caption{Overall Performance Comparison with Baseline Methods in Simulation.}
	\label{tab2}
	\begin{tabular}{l|cccc|cccc}
		\Xhline{1.2pt}
		\multirow{2}{*}{\centering Methods} 
		& \multicolumn{4}{c|}{Task Success Rate $\eta_p$ $\uparrow$} 
		& \multicolumn{4}{c}{Per-step Interference time $T_s$ $\downarrow$} \\ \cline{2-9} 
		& Insert Cylinder & Pinch Pen & Grasp Handle & Move Card 
		& Insert Cylinder & Pinch Pen & Grasp Handle & Move Card \\ 
		\hline
		ACT+3D\cite{zhao2023learning}   &23\%  &25\%  &46\%  &34\%  &4.5ms  &4.6ms  &4.5ms  &4.3ms  \\
		DP3\cite{ze2024dp3}      &83\%  &77\%  &80\%  &53\%  &29.1ms  &31.5ms  &29.8ms  &29.6ms  \\
		ManiCM\cite{lu2024manicm}   &24\%  &22\%  &54\%  &45\%  &4.4ms  &4.8ms  &5.2ms  &5.5ms  \\
		IDP3\cite{ze2025generalizable}     &80\%  &68\%  &73\%  &86\%  &31.6ms  &32ms  &31.7ms  &32.4ms  \\
		Flow Policy\cite{fang2025flow}     &25\%  &48\%  &43\%  &47\%  &3.2ms  &\textbf{3.7ms}  &\textbf{3.3ms}  &\textbf{3.6ms}  \\
		ResiP\cite{ankile2025imitation}    &85\%  &79\%  &80\%  &87\%  &29.3ms  &32.5ms  &31.9ms  &30.2ms  \\
		\textbf{FAR-DexRes (Ours)}     &\textbf{93\%}  &\textbf{83\%}  &\textbf{88\%}  &\textbf{95\%}  &\textbf{3.0ms}  &4.3ms  &3.8ms  &4.3ms  \\ 
		\Xhline{1.2pt}
	\end{tabular}
\end{table*}

\subsubsection{Baselines}
In the data generation stage, we compare FAR-DexGen with MimicGen\cite{mandlekar2023mimicgen} and DemoGen\cite{xue2025demogen} to evaluate differences in synthesis efficiency and data quality. In the policy execution stage, FAR-DexRes is benchmarked against several mainstream baselines, including ACT+3D\cite{zhao2023learning}, DP3\cite{ze2024dp3}, ManiCM\cite{lu2024manicm}, IDP3\cite{ze2025generalizable}, FlowPolicy\cite{fang2025flow}, and ResiP\cite{ankile2025imitation}. The first five methods belong to end-to-end imitation learning paradigms, while ResiP combines imitation learning with reinforcement learning refinement. To ensure fairness in comparison, we implement ResiP with its base policy replaced by DP3 instead of DP.

\subsubsection{Evaluation metrics} At the data generation level, we evaluate efficiency and quality using two metrics: time per trajectory generation $T_g$ and generation quality $\eta_g$. At the policy execution level, we adopt task success rate $\eta_p$ and per-step inference time $T_s$ as evaluation criteria.

\subsection{Data Generation Performance}

In the data generation evaluation, we compare FAR-DexGen with MimicGen and DemoGen. For the Insert-Cylinder task, 100 trajectories are first generated, with the average generation time per trajectory $T_g$ serving as the efficiency metric. To evaluate the data generation quality, we train a unified DP3 baseline individually on the datasets produced by each candidate method. The task success rates of these resulting models are recorded as an indirect proxy to quantify the quality of the underlying data generation ($\eta_g$).

As shown in Table I, MimicGen achieves the fastest generation speed, followed by DemoGen, while FAR-DexGen requires slightly more time due to point cloud alignment and joint angle conversion. However, the gap with the fastest method is only 2 ms, demonstrating strong competitiveness. In terms of quality, by incorporating fine-grained hand–object interaction details and physical constraints, FAR-DexGen significantly improves data generation quality, yielding success rates 19.6\% and 13.4\% higher than MimicGen and DemoGen, respectively, thereby demonstrating clear advantages in both efficiency and quality.

\begin{figure}[]
	\centering
	\includegraphics[width=0.42\textwidth]{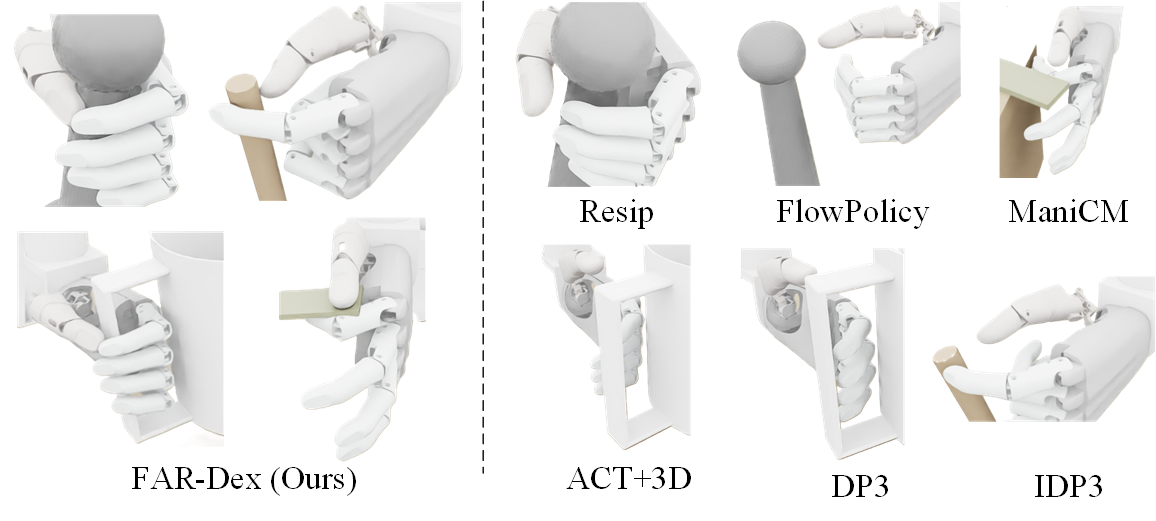}
	\captionsetup{font=small,skip=8pt,belowskip=0pt}
	\caption{Visualization of object interactions in simulation, compared with typical failure cases from existing methods}
	\label{fig:5}
\end{figure}

\subsection{Policy Performance in Simulation}
In simulation, we construct four dexterous manipulation tasks, and evaluate each method by executing 100 trials per task in simulation. At each reset, the planar position of the object is randomly perturbed by 1 cm. The evaluation metrics include task success rate $\eta_p$ and per-step inference speed $T_s$.

\subsubsection{Success Rate Comparison} 
As shown in the left panel of Table II, FAR-DexRes achieves over 83\% success across all four tasks, with Insert and Move reaching 93\% and 95\%, respectively, representing an average improvement of 7\% over the best baseline, ResiP. Fig. 5 further illustrates its superior stability and coherence during hand–object contact. In comparison, ACT+3D struggles to generate smooth actions, DP3 and IDP3 lack fine-grained coordination, ManiCM and FlowPolicy often suffer from premature finger closure in high-dimensional action spaces, and ResiP, while benefiting from RL-based refinement to improve interaction details, remains limited by its uniform residual scaling, which cannot flexibly generate multi-DoF actions.

By contrast, FAR-DexRes incorporates trajectory segmentation and temporal information, and outputs adaptive residual weights aligned with the dimensionality of the action space, enabling dynamic coordination between the robotic arm and multi-fingered hand across different task phases. This mechanism ensures stability in long-horizon complex tasks while achieving greater precision in fine-grained manipulation, leading to superior performance across all four tasks.
\subsubsection{Inference Speed Comparison} 
The right panel of Table II presents the per-step inference time of different methods across the four tasks. DP3, IDP3, and ResiP require multi-step denoising iterations with large conditional U-Nets\cite{ho2020denoising}, resulting in an average of about 30 ms per step. ACT+3D incurs its primary overhead from the Transformer encoder–decoder structure, remaining stable at around 5 ms. ManiCM and FlowPolicy compress multi-step denoising into single-step inference through consistency distillation\cite{song2023consistency} and flow matching\cite{lipman2022flow}, achieving the fastest speed in some tasks but with limited overall success rates.

In contrast, FAR-DexRes employs consistency distillation together with efficient convolutional feature extraction and a lightweight spatio-temporal adaptive module, keeping overhead below 0.03 ms, which is negligible. Although not the fastest in every task, it consistently maintains an inference time around 3.8 ms while achieving the highest success rates across all four tasks, demonstrating the best balance between accuracy and speed. This level of latency also meets real-world deployment requirements, ensuring reliable control.

\begin{table}[]
	\centering
	\renewcommand{\arraystretch}{1.2}
	\captionsetup{font=small}
	\caption{Ablation Study of the RL Components.}
	\label{tab:ablation}
	\begin{tabular}{l|cccc}
		\Xhline{1.2pt}
		Ablation      
		& Insert
		& Pinch 
		& Grasp 
		& Move \\ 
		\Xhline{0.8pt}
		w/o RL         & 86\% & 65\% & 72\% & 73\% \\
		w/o Traj-Embed. & 82\% & 58\% & 69\% & 80\% \\
		w/o Obs-Feat.   & 86\% & 78\% & 74\% & 82\% \\
		w/o Time-steps & 87\% & 75\% & 70\% & 86\% \\
		\textbf{FAR-DexRes (Ours)}  
		& \textbf{93\%} & \textbf{83\%} & \textbf{88\%} & \textbf{95\%} \\ 
		\Xhline{1.2pt}
	\end{tabular}
\end{table}

\subsection{Ablation Study}

To further assess the contribution of RL-based refinement, we conduct an ablation study on its individual components. As shown in Table III, removing RL refinement leads to an overall drop in success rates across all four tasks, indicating that RL refinement effectively supplements fine-grained interaction information.

Additional ablations highlight the critical role of trajectory segmentation and temporal modeling in dexterous manipulation. For two-finger coordination tasks such as Pinch Pen and Move Card, removing trajectory embedding reduces success rates by 25\% and 15\%, respectively. For Insert Cylinder and Grasp Handle tasks, incorporating observation features improves success rates by 7\% and 14\%. Notably, in the absence of trajectory embedding, even with RL refinement, performance falls below that of a direct visuomotor policy, suggesting that sparse rewards alone are insufficient to support fine-grained interactions.

\begin{figure}[]
	\centering
	\includegraphics[width=0.45\textwidth]{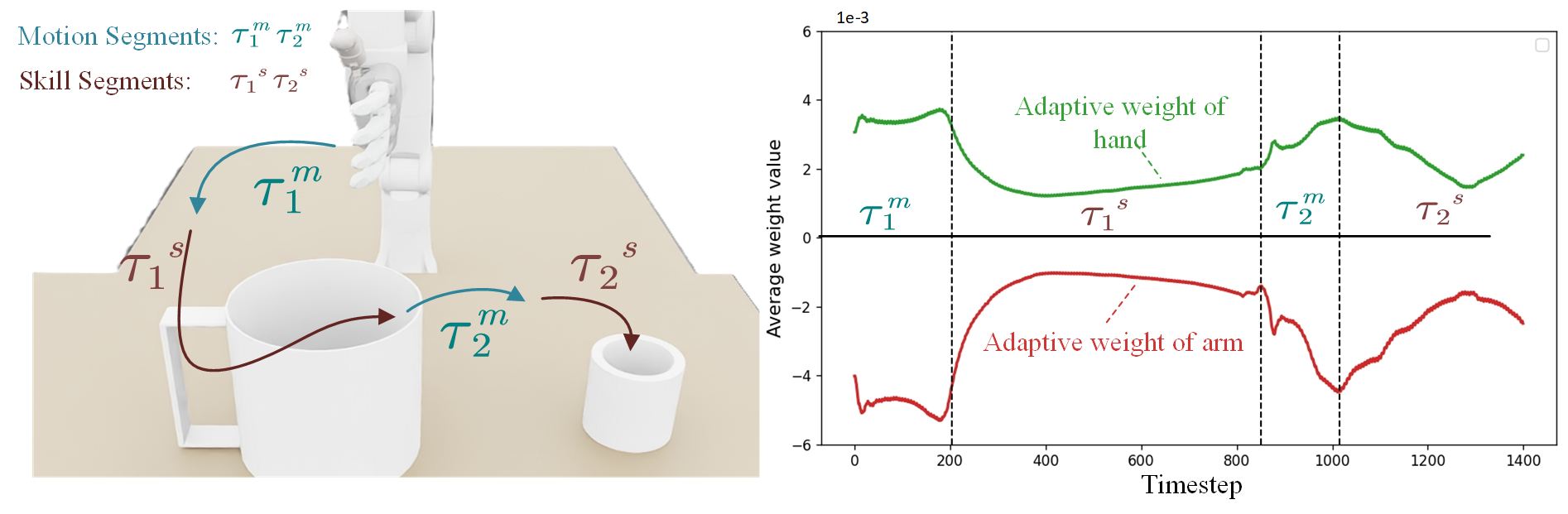}
	\captionsetup{font=small,skip=8pt,belowskip=0pt}
	\caption{Residual Weight Variation Curves, where the Hand and Arm Weights Dynamically Change across Different Trajectory Segments.}
	\label{fig:6}
\end{figure}

\begin{figure}[]
	\centering
	\includegraphics[width=0.42\textwidth]{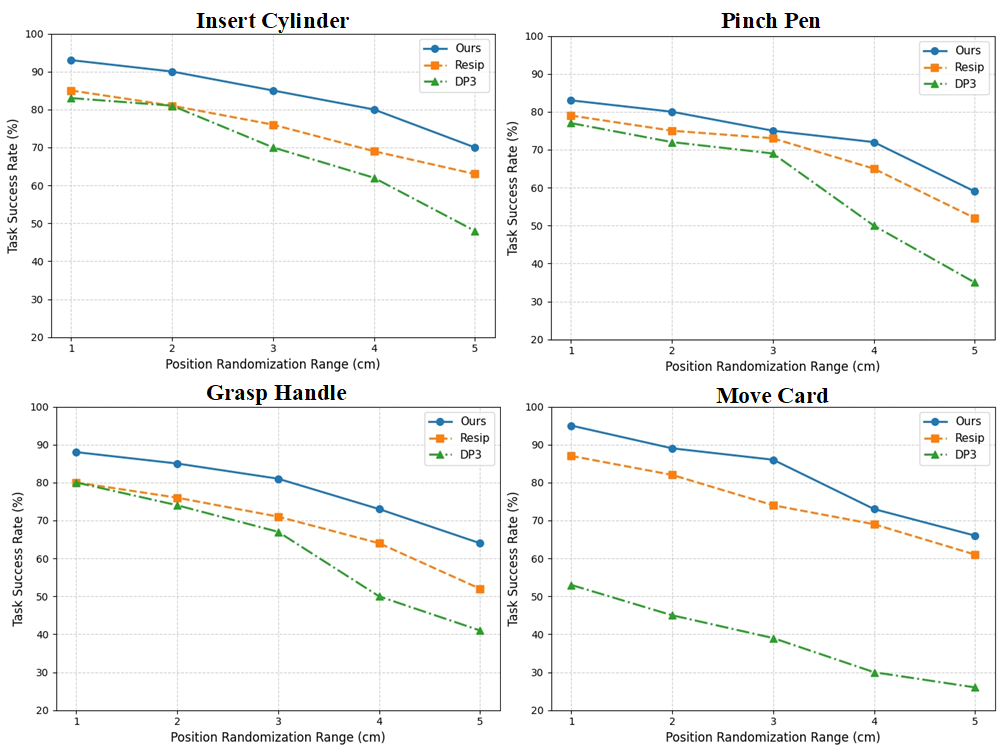}
	\captionsetup{font=small,skip=8pt,belowskip=0pt}
	\caption{Position Generalization Success Rate Curves. As the degree of randomization increases, we record the task success rates of FAR-DexRes(Ours), DP3\cite{ze2024dp3}, and ResiP\cite{ankile2025imitation}.}
	\label{fig:7}
\end{figure}

\subsection{Robustness Analysis}
\subsubsection{Residual Weight Analysis}
The previous ablation study confirmed the critical role of trajectory embedding in dexterous manipulation. To further investigate their underlying mechanism, we analyze the temporal evolution of adaptive residual weights in the Grasp Handle task.  Specifically, the residual weights $\sigma_t \in \mathbb{R}^{1\times17}$ are divided into arm weights $\sigma^{\text{arm}}_t \in \mathbb{R}^{1\times7}$ and hand weights $\sigma^{\text{hand}}_t \in \mathbb{R}^{1\times10}$, and their averages are recorded over time, as illustrated in Fig. 6.

The results indicate that $\bar{\sigma}^{\text{arm}}_t$ and $\bar{\sigma}^{\text{hand}}_t$ apply distinct adjustments to the robotic arm and dexterous hand at the same time step. During the initial motion phase, the residual policy dominates: $\bar{\sigma}^{\text{arm}}_t$ becomes negative to constrain deviations in arm motion, while $\bar{\sigma}^{\text{hand}}_t$ is positive to guide the hand into a pre-grasp state. As the interaction skill phase begins, both weights gradually approach zero, indicating that the base policy assumes control while the residual policy provides only fine-grained corrections to ensure stable and precise contact. This dynamic pattern is consistently reproduced in subsequent motion and skill phases.

These findings demonstrate that residual weights with trajectory embeddings allow flexible allocation of arm–hand coordination across different stages: actively correcting deviations in motion segments and maintaining fine-grained control in skill segments. This mechanism substantially enhances both the stability and precision of dexterous manipulation.

\subsubsection{Positional Generalization Analysis}
However, relying solely on the dynamic adjustment of trajectory segments is insufficient to demonstrate robustness in diverse scenarios. To further validate the positional generalization capability of FAR-DexRes, we introduced random perturbations of 1–5 cm in the $x$–$y$ directions to the initial object poses across four tasks, generating 100 trajectories for testing under each setting. As shown in Fig. 7, FAR-DexRes maintained success rates above 55\% even under extreme perturbations of 5 cm, significantly outperforming ResiP and DP3. These results indicate that, by leveraging multi-steps trajectory embedding and observation feature extraction, FAR-DexRes can establish correct correspondences even when test conditions deviate from the training distribution, thereby exhibiting stronger positional generalization and robustness.

\begin{table}[]
	\centering
	\renewcommand{\arraystretch}{1.2}
	\captionsetup{font=small}
	\caption{Performance of FAR-DexRes in the Real World.}
	\label{tab:comparison}
	\begin{tabular}{l|cccc}
		\Xhline{1.2pt}
		Method
		& Insert
		& Pinch 
		& Grasp 
		& Move \\ 
		\Xhline{0.8pt}
		ACT+3D\cite{zhao2023learning}       & 20\% & 0\%  & 40\% & 25\% \\
		DP3\cite{ze2024dp3}          & 65\% & 70\% & 70\% & 45\% \\
		ManiCM\cite{lu2024manicm}       & 20\% & 0\%  & 35\% & 40\% \\
		IDP3\cite{ze2025generalizable}         & 75\% & 65\% & 65\% & 80\% \\
		FlowPolicy\cite{fang2025flow}   & 25\% & 10\% & 35\% & 40\% \\
		ResiP\cite{ankile2025imitation}        & 80\% & 70\% & 75\% & 80\% \\
		\textbf{FAR-DexRes (Ours)} 
		& \textbf{85\%} & \textbf{80\%} & \textbf{80\%} & \textbf{90\%} \\ 
		\Xhline{1.2pt}
	\end{tabular}
\end{table}

\begin{figure}[]
	\centering
	\includegraphics[width=0.45\textwidth]{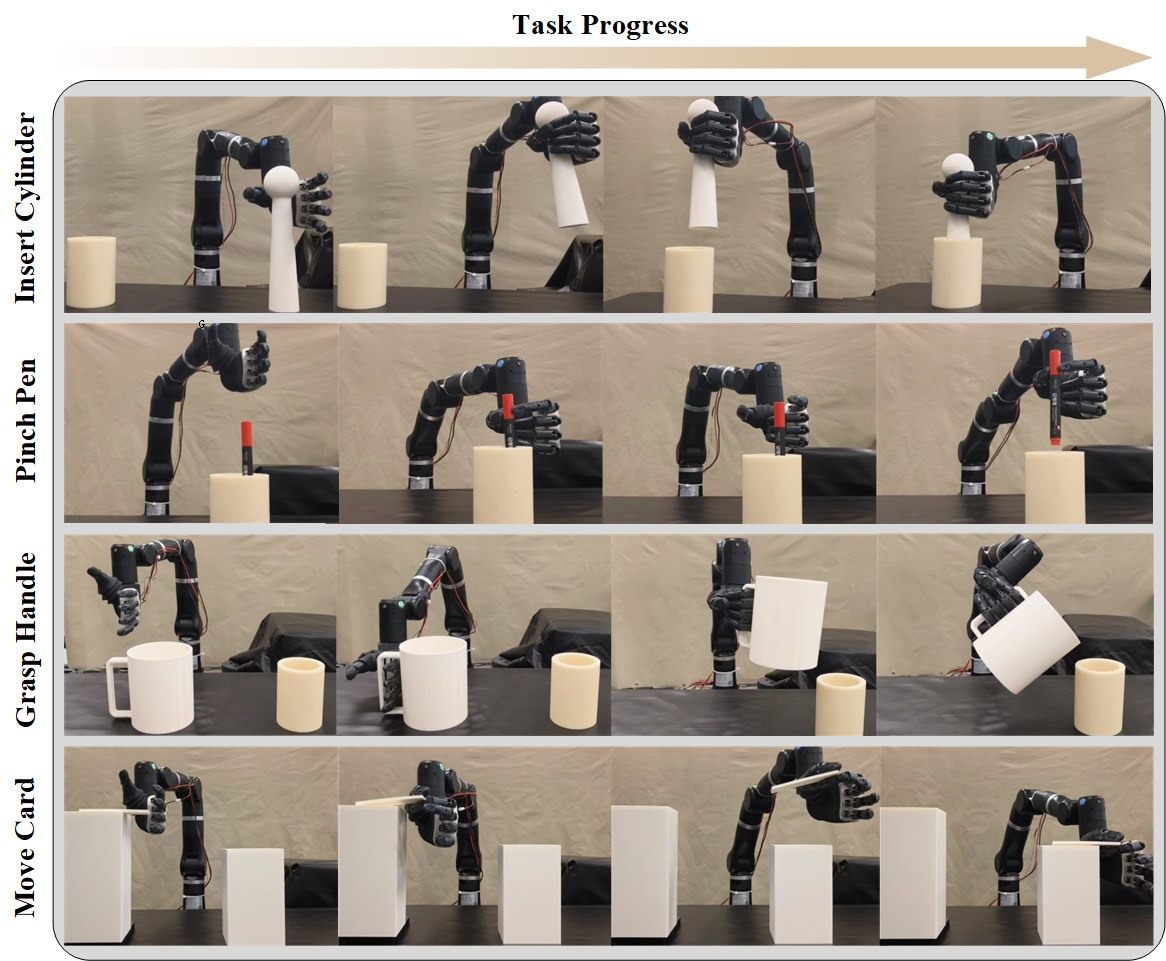}
	\captionsetup{font=small,skip=8pt,belowskip=0pt}
	\caption{Real-World Validation}
	\label{fig:8}
\end{figure}

\subsection{Real-World Validation}
After completing the simulation experiments, we deployed the trained policies to real-world settings and conducted 20 trials for each task. Due to sensing delays and environmental disturbances, all methods exhibited slightly lower success rates in reality compared to simulation. As shown in Table IV, our approach achieved success rates exceeding 80\% across all tasks, consistently outperforming the baselines, with improvements of 10\% over ResiP in the Pinch and Move tasks. Furthermore, as illustrated in Fig. 8, our method demonstrated more stable and precise performance in both five-finger grasping and two-finger pinching scenarios.

\section{Conclusion and Limitation}
This work presents FAR-Dex, a framework that integrates efficient data augmentation with adaptive residual refinement for dexterous manipulation. Experimental results show that FAR-Dex outperforms existing methods in both data quality and task success rates. Further analysis verifies that FAR-Dex adaptively adjusts individual arm–hand action components by introducing spatio-temporal features, enabling higher precision control and positional generalization. Real-world experiments further validate its potential for handling complex multi-finger manipulation tasks under few-shot conditions.

Despite its success, FAR-Dex faces high simulation costs and the inherent limitations of domain randomization in bridging the sim-to-real gap. Future work will focus on integrating 3D rendering and efficient data collection to reduce transfer errors. Additionally, incorporating force and tactile sensing will be essential to enhance precision and robustness in complex manipulation.

\bibliographystyle{IEEEtran}   
\bibliography{refs}            
\end{document}